\documentclass[letterpaper, 10 pt, conference]{ieeeconf}  
\usepackage{booktabs}

\IEEEoverridecommandlockouts                              

\usepackage{graphics} 
\usepackage{epsfig} 
\usepackage{mathptmx} 
\usepackage{times} 
\usepackage{amsmath} 
\usepackage{amssymb}  
\usepackage{bm}       
\usepackage{algorithm}
\usepackage{algpseudocode}
\usepackage[table]{xcolor} 
\usepackage{float} 
\usepackage{graphicx} 
\usepackage{subcaption} 
\usepackage{wrapfig}
\usepackage{cite} 
\usepackage{lipsum}  
\usepackage{xspace}
\usepackage[colorlinks=true, urlcolor=blue, linkcolor=red]{hyperref}
\usepackage{url}
\usepackage{listings}
\usepackage{xcolor}
\usepackage[draft]{changes}
\usepackage{flushend}

\title{\LARGE \bf Safe Navigation of Bipedal Robots via \\Koopman Operator-Based Model Predictive Control}

\author{Jeonghwan Kim, Yunhai Han, Harish Ravichandar, Sehoon Ha
\thanks{JK, YH, HR, and SH are with the School of Interactive Computing at the Georgia Institute of Technology, Atlanta, USA. {\tt\small \{jkim3662\}@gatech.edu}}
}

\newcommand{\cmt}[1]{}

\long\def\ignorethis#1{}

\newcommand{\pctab}{\hspace{0.2in}}

\begin{document}

\maketitle
\thispagestyle{empty}
\pagestyle{empty}

\begin{abstract}
Nonlinearity in dynamics has long been a major challenge in robotics, often causing significant performance degradation in existing control algorithms.
For example, the navigation of bipedal robots can exhibit nonlinear behaviors even under simple velocity commands, as their actual dynamics are governed by complex whole-body movements and discrete contacts.
In this work, we propose a safe navigation framework inspired by Koopman operator theory.
We first train a low-level locomotion policy using deep reinforcement learning, and then capture its low-frequency, base-level dynamics by learning linearized dynamics in a high-dimensional lifted space.
Then, our model-predictive controller (MPC) efficiently optimizes control signals via a standard quadratic objective and the linear dynamics constraint in the lifted space.
We demonstrate that the Koopman model more accurately predicts bipedal robot trajectories than baseline approaches.
We also show that the proposed navigation framework achieves improved safety with better success rates in dense environments with narrow passages.
\end{abstract}

\section{INTRODUCTION}
\enlargethispage{1.4\baselineskip}
Nonlinear dynamics have long been a fundamental challenge in robotics, which often lead to significant difficulties in modeling, prediction, and control.
Legged robot navigation is one of the examples because even simple velocity commands can induce highly nonlinear behaviors due to their complex hybrid dynamics.
Researchers have explored both model-based and reinforcement learning-based approaches.
Model-based methods often integrate obstacle avoidance constraints with model predictive control (MPC), leveraging explicit dynamic models for safe navigation~\cite{tonneau2018efficient, di2018dynamic, wellhausen2021rough, asselmeier2024hierarchical}.
On the other hand, reinforcement learning-based frameworks have achieved impressive results in legged locomotion and navigation, either by training high-level policies~\cite{merel2018hierarchical, hoeller2021learning, hoeller2024anymal} or developing end-to-end policies~\cite{yang2023neural, zhuang2023robot, cheng2024extreme}.
While both lines of research have been successful, model-based approaches are constrained by the complexity of accurate dynamics modeling and the increased computation cost, whereas reinforcement learning-based methods demand extensive training data and tend to show degraded performance in unseen environments.
Poor generalization raises critical safety concerns because unpredictable navigation in unseen environments often leads to collisions or falls.

Researchers have explored various algorithms that combine the strengths of both model-based and reinforcement learning-based approaches~\cite{xie2022glide,jung2025ppf}.
A notable example is the work of Li et al.~\cite{li2022bridging}, which identifies low-dimensional linear models of RL-based closed-loop dynamics in legged robots via system identification.
Their results demonstrate that the learned models can successfully enable safety-critical optimal control, such as MPC with control barrier functions, on physical systems.
However, this method learns low-dimensional, component-wise dynamics, which may not sufficiently capture the nonlinearity of high-level behaviors in more challenging scenarios.

\begin{figure}
    \centering
    \includegraphics[width=0.48\textwidth]{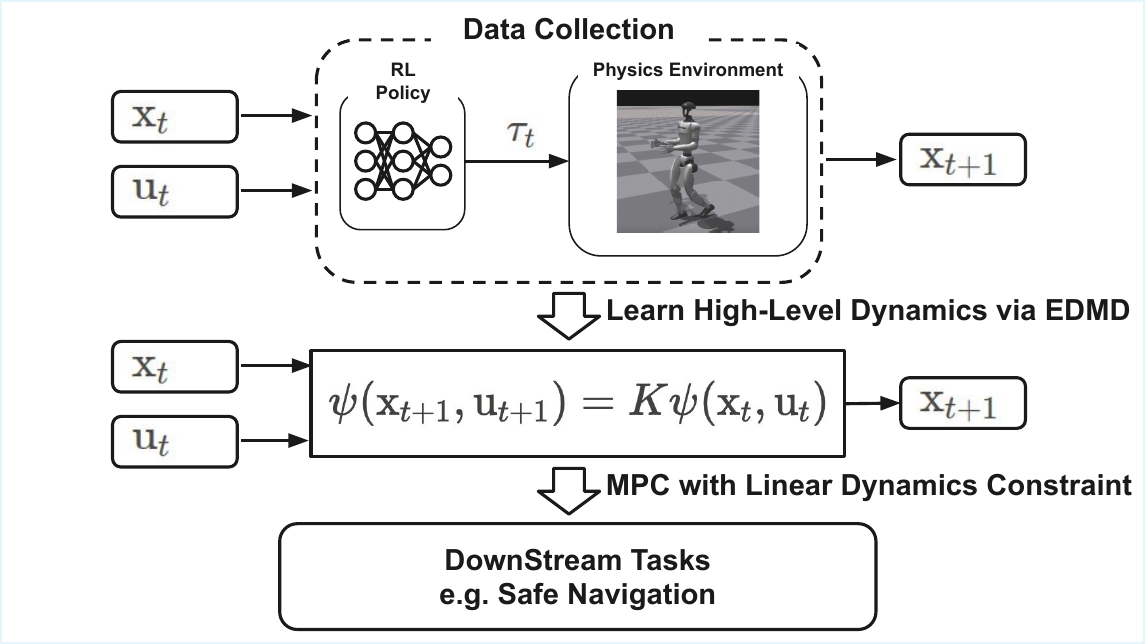}
    \caption{
Overview of the proposed framework. Our approach (1) trains a robust locomotion controller for the humanoid robot, (2) identifies the high-level dynamics via Koopman operator theory for linearization, and (3) integrates the learned model into an MPC-based controller for safe navigation.
}
    \label{fig: overview}
\end{figure}

\begin{figure*}[t]
    \centering
    \begin{subfigure}[b]{0.32\textwidth}
        \centering
    \includegraphics[width=\textwidth]{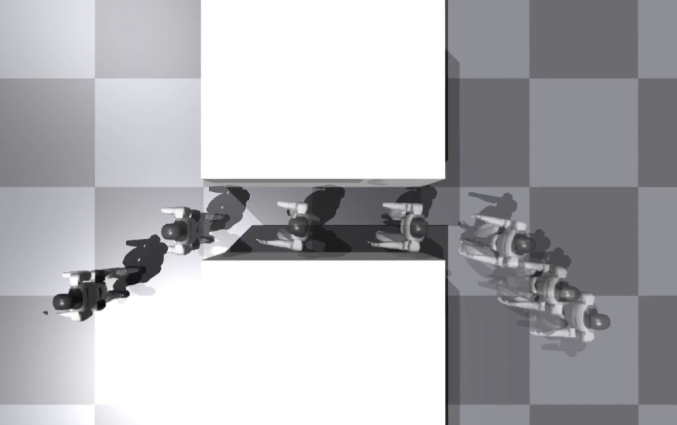}
        \caption{Narrow Corridor}
    \end{subfigure}
    \hfill
    \begin{subfigure}[b]{0.32\textwidth}
        \centering
    \includegraphics[width=\textwidth]{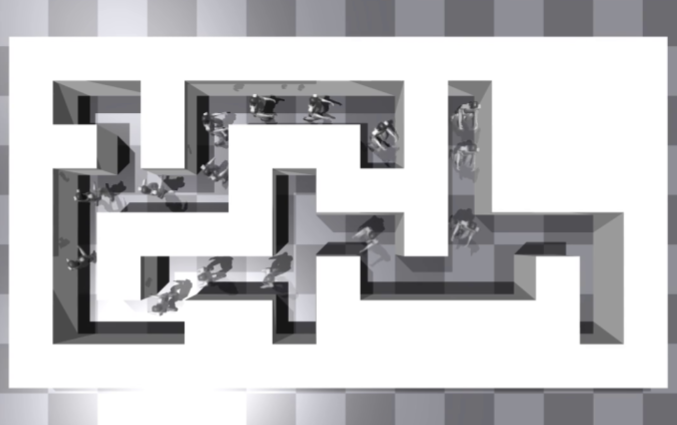}
        \caption{Maze}
    \end{subfigure}
    \hfill
    \begin{subfigure}[b]{0.32\textwidth}
        \centering
    \includegraphics[width=\textwidth]{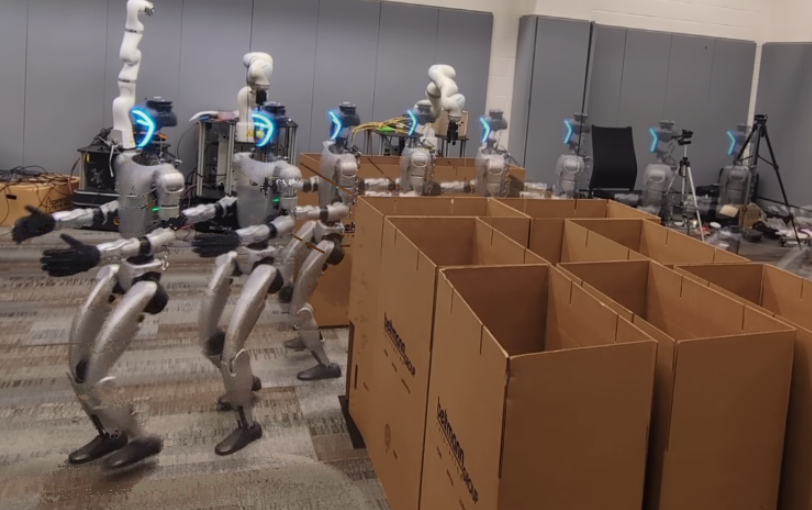}
        \caption{Hardware Validation}
    \end{subfigure}
    \caption{Demonstrations of safe navigation: (a) narrow corridor traversal, (b) maze navigation with dense obstacles, and (c) hardware validation on the physical Unitree G1 humanoid.}
    \label{fig:safe_navigation}
\vspace{-12pt}
\end{figure*}

In this work, we leverage Koopman operator theory to represent the nonlinear dynamics of bipedal robots as linear transitions in a high-dimensional lifted space~\cite{schmid2010dynamic, Koopman1931Koopman}.
We first train a low-level locomotion policy via deep reinforcement learning and then capture its low-frequency, base-level dynamics using Extended Dynamic Mode Decomposition (EDMD)~\cite{williams2015data}.
Unlike existing approaches that rely on simplified analytic models or black-box neural networks, our framework provides a simple yet accurate representation of dynamics.
Building on this model, we formulate an MPC framework using a standard quadratic objective and a linear dynamics constraint, enabling both computational efficiency and reliable performance for safety-critical navigation tasks.

We conducted experiments to evaluate our framework in highly constrained environments, such as narrow mazes and corridors, where minimal clearance makes navigation particularly challenging and collisions often lead to falling or getting stuck.
Our results show that the Koopman dynamics model with phase augmentation significantly outperforms baseline linear and MLP models in long-term prediction. Specifically, for a 12-step (6-second) rollout, the Koopman model reduced positional error by 50\% (0.188 m) compared to the linear baselines~\cite{li2022bridging} (0.374 m).
When integrated with MPC for safe navigation, this superior predictive accuracy translated into substantial gains in reliability.
While linear baselines struggle with the frequent turning maneuvers required in complex layouts, our Koopman-based framework achieved a 96\% overall success rate across four maps of varying difficulty compared to the linear baseline of 86\%.
Our Koopman model was able to reduce the peak violation depth by $47.5\%$ compared to the baselines.
Finally, we successfully deploy the proposed framework on physical hardware to demonstrate its capability for safe navigation.
The main contributions of this work are:
\begin{itemize}  
    \item We propose a Koopman-based safe navigation control framework for bipedal robots, which leverages learned linear Koopman dynamics and model-predictive control.  
    \item We provide a comprehensive evaluation of various forward dynamics models on bipedal robots, including an integrator, component-wise linear model, a linear model, and Koopman dynamics models with different lifting functions including phase augmentation.
    \item We demonstrate improved safety of the proposed framework compared to the baselines.  
\end{itemize}

\section{Related Works}
\label{sec:related}
\subsection{Safe Legged Locomotion}
It is important to ensure safety during legged robots' dynamic motions with underactuated bases.
Previously, safety has been widely discussed within the context of model-based control~\cite{winkler2018gait, di2018dynamic}.
Recent advances in learning-based locomotion frameworks~\cite{haarnoja2018soft, rudin2022learning} have demonstrated impressive robustness, but there are only limited works that discuss the interpretability~\cite{glanois2024survey} and guarantee of safety~\cite{he2024agile}.
A common approach in this context is to decouple a low-level locomotion controller and a high-level navigation policy, where a high-level policy adopts conventional methods such as model predictive control and control barrier functions~\cite{ames2019control, grandia2021multi, choi2020reinforcement}.
Li et al.~\cite{li2022bridging} propose a method that leverages low-dimensional linear models identified from bipedal locomotion policies trained via reinforcement learning.
Liao et al.~\cite{liao2023walking} model robots and obstacles as polytopes~\cite{thirugnanam2022safety} to enable narrow space navigation.
However, they require access to a low-level whole-body controller, which is not available in many real-world scenarios.
Instead of using analytic formulations, there exist works to model low-level behaviors using neural networks~\cite{kim2023armp, kim2022learning}.
However, the learned dynamics are often highly nonlinear, which makes it infeasible to adopt conventional optimization methods and requires additional techniques, such as an informed trajectory sampler~\cite{kim2022learning}.

\subsection{Koopman Operator for Robotics}
The Koopman operator theory~\cite{koopman1932dynamical}, which offers a linear perspective on nonlinear dynamical systems, has gained growing interest in robotics for its potential to simplify complex dynamical systems~\cite{abraham2017model} and high sample efficiency~\cite{han2023utility}.
It has been successfully utilized in various domains such as spherical robots~\cite{abraham2017model}, drones~\cite{folkestad2022koopnet, folkestad2021koopman, folkestad2020data}, and dexterous manipulation~\cite{han2023utility, han2024learning, chen2024korol}.
Leveraging their linearity, Koopman operator methods have widely been combined with model predictive control to obtain desired command tracking~\cite{abraham2017model, korda2018linear}, enforce better stability, or improve safety.
The Koopman operator has also proven useful in learning the reference dynamics from trajectory data, which demonstrates superior sample efficiency compared to deep neural networks in dexterous manipulation~\cite {han2023utility}.
In this work, we learn the high-level dynamics of the velocity-tracking policy and leverage a model predictive control framework to obtain safe commands to navigate within environments with dense obstacles.

\section{Preliminaries: Learning Koopman Dynamics}
\label{sec:preliminary}
We briefly review Koopman operator theory~\cite{Koopman1931Koopman}, which provides a principled framework for representing nonlinear dynamics as linear evolution in a lifted observable space.

\subsubsection{Koopman Representation}
Consider a nonlinear system $\bm{x}_{t+1} = F(\bm{x}_t)$,
where $\bm{x}_t \in \mathcal{X} \subset \mathbb{R}^n$ denotes the state at time $t$, and $F(\cdot):\mathbb{R}^n \rightarrow \mathbb{R}^n$ is a nonlinear transition function.
The key idea of Koopman operator theory is to introduce a set of \textit{observable} functions $g:\mathcal{X} \rightarrow \mathbb{R}$ drawn from an infinite-dimensional function space, and to study their evolution rather than that of the state directly.
The \textit{Koopman operator} $\mathcal{K}$ is then defined as the composition:
$$(\mathcal{K}g)(\bm{x}_t) = g(F(\bm{x}_t)) = g(\bm{x}_{t+1}).$$
Although $F$ may be nonlinear, $\mathcal{K}$ acts \textit{linearly} on the space of observables:
$g(\bm{x}_{t+1}) = \mathcal{K}g(\bm{x}_t)$.
This linearity holds exactly but requires an \textit{infinite-dimensional} operator, making direct computation infeasible.

In practice, we focus directly on the controlled setting relevant to our framework.
For a controlled system $\bm{x}_{t+1} = F(\bm{x}_t, \bm{u}_t)$ with $\bm{u}_t \in \mathcal{U} \subset \mathbb{R}^m$, a common strategy~\cite{abraham2017model} is to choose a finite-dimensional \textit{lifting function} $\phi:\mathcal{X} \rightarrow \mathbb{R}^p$ and form the full observable by concatenating the lifted state with the control input:
$$\psi(\bm{x}_t, \bm{u}_t) = [\phi(\bm{x}_t);\, \bm{u}_t] \in \mathbb{R}^{p+m}.$$
The Koopman operator is then approximated by a finite matrix $K \in \mathbb{R}^{(p+m)\times(p+m)}$ acting on this joint observable:
$$
[\phi(\bm{x}_{t+1});\, \bm{u}_{t+1}] = K[\phi(\bm{x}_t);\, \bm{u}_t] + r(\bm{x}_t, \bm{u}_t),
$$
where $r(\bm{x}_t, \bm{u}_t) \in \mathbb{R}^{p+m}$ is the truncation residual.

\noindent Partitioning $K$ block-wise as
$K = \bigl[\begin{smallmatrix} A & B \\ C & D \end{smallmatrix}\bigr]$
and retaining only the rows governing state evolution, we obtain:
$$
 \phi(\bm{x}_{t+1}) =  A\phi(\bm{x}_t) + B\bm{u}_t + r(\bm{x}_t, \bm{u}_t),
$$
where $A \in \mathbb{R}^{p \times p}$ and $B \in \mathbb{R}^{p \times m}$.

\subsubsection{Learning Koopman Dynamics}
The operators $A$ and $B$ are learned from a dataset $\mathcal{D} = [(\bm{x}_1, \bm{u}_1), \cdots, (\bm{x}_T, \bm{u}_T)]$ of state trajectories and control inputs. Given the choice of lifting function $\phi(\cdot)$, we find $A$ and $B$ by minimizing the prediction residual:
\begin{equation}
    \label{equ:dynamics_loss}
J(K) = \sum_{t=1}^{T-1} \Vert r(\bm{x}_t, \bm{u}_t) \Vert ^2 = \sum_{t=1}^{T-1} \Vert [\phi(\bm{x}_{t+1}); \bm{u}_{t+1}] - K[\phi(\bm{x}_t); \bm{u}_t] \Vert ^2.
\end{equation}
Because this is a least-squares problem, we can obtain an analytical solution. We define $X$ and $Y$ as:
\begin{align*}
    X &= \frac{1}{T-1}\sum_{t=1}^{T-1}[\phi(\bm{x}_{t+1}); \bm{u}_{t+1}] \otimes [\phi(\bm{x}_t); \bm{u}_t] \\
    Y &= \frac{1}{T-1}\sum_{t=1}^{T-1}[\phi(\bm{x}_t); \bm{u}_t] \otimes [\phi(\bm{x}_t); \bm{u}_t],
\end{align*}
where $\otimes$ denotes the outer product. Then, we can take the Moore-Penrose pseudoinverse $K = XY^\dagger$ as suggested in~\cite{schmid2010dynamic, williams2015data}.
The specific form of the lifting function $\phi(\cdot)$ used in this work is discussed in Section~\ref{sec:exp}.

\section{Safe Navigation with Koopman Forward Dynamics}
\label{sec:framework}
This section presents a safe navigation framework that leverages the learned Koopman dynamics.
Model-predictive control (MPC) has been widely adopted as a theoretical backbone for safe navigation.
However, the dynamics of a bipedal robot are difficult to analyze because it is a closed-loop system with a nonlinear control policy and hybrid contact dynamics, which can degrade the performance of the existing safety-aware control algorithms.

Instead, we model the robot's closed-loop forward dynamics as a linear evolution in a lifted observable space using the Koopman operator.
This linear structure integrates naturally with MPC, enabling efficient and safety-aware navigation planning.

\subsection{Koopman Forward Dynamics} \label{subsec:learning_dynamics}

\label{sec:KoopmanMPC}
\begin{figure}[t]
    \centering
    \includegraphics[width=0.45\textwidth]{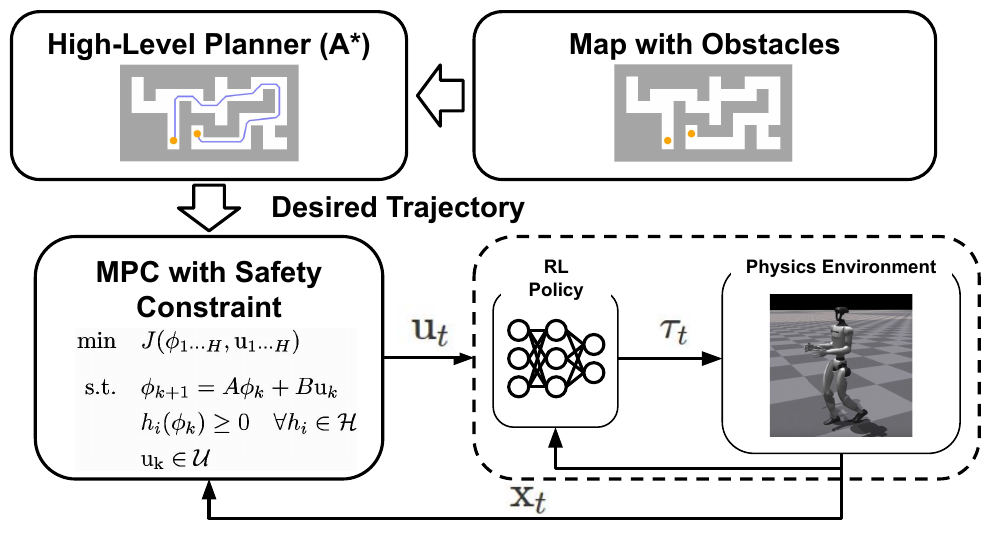}
    \caption{Overview of the navigation framework.}
    \label{fig:navigation}
\vspace{-5pt}
\end{figure}

Our main objective in this section is to learn the state transition function $\bm{x}_{t+1} = F(\bm{x}_t, \bm{u}_t)$, where $F(\cdot): \mathcal{X} \times \mathcal{U} \to \mathcal{X}$ is a high-level dynamics that governs the closed-loop forward dynamics of the legged locomotion. 

We define the high-level state as $\bm{x} = [p_x, p_y, \theta, v_x, v_y, \omega] \in \mathbb{R}^6$, where $p_x, p_y$ are the planar position, $\theta$ is the heading angle, and $v_x, v_y, \omega$ are the linear and angular velocities, respectively.
The control input is $\bm{u} = [\hat{v}_x, \hat{v}_y, \hat{\omega}] \in \mathbb{R}^3$, representing the commanded velocities expressed in the robot's local frame.

One key challenge in learning $F$ is that it can be arbitrarily nonlinear.
One potential way in machine learning is to utilize a universal function approximator, a multi-layer perceptron (MLP), to model the input-output mapping of the function $F$ minimizing the $L_2$ loss: $||\bm{x}_{t+1} -f_{\text{MLP}}(\bm{x}_t, \bm{u}_t)||_2$ $\forall t$.
However, utilizing $\text{MLP}$ dynamics in the MPC framework causes practical challenges, which makes MLP-learned dynamics impractical for general usage.
For instance, Direct Multiple Shooting formulation~\cite{bock1984multiple, giftthaler2018family, thirugnanam2022safety}, one of the widely used techniques in solving MPC, requires the dynamics constraint containing $\text{MLP}$ to match for every time step, making the optimization extremely nonconvex.
As a result, nonlinear solvers often fail to solve the optimization with MLP dynamics constraints. 

Instead, we use the Koopman operator's linear evolution to represent dynamics in the lifted space, which makes the dynamics constraint in the MPC framework linear. As a result, our approach offers the possibility of adopting linear MPC with control barrier functions. In this work, we use the Koopman forward dynamics as described in Section~\ref{sec:preliminary}: $\phi(\bm{x}_{t+1}) = A\phi(\bm{x}_t) + B\bm{u}_t$, 
where the state-lifting, $\phi(\cdot) : \mathbb{R}^n \to \mathbb{R}^{n'}$, $n' \geq n$, is a vector-valued lifting function. Inspired by~\cite{han2023utility}, the first $n$ elements of $\phi(\bm{x}_t)$ are $\bm{x}_t$ so that we can easily retrieve the original states, which are used for computing obstacle-avoidance constraints.
Note that in the case of $n' = n$, $\phi(\bm{x})$ becomes the same as $\bm{x}$, and the solutions $A$ and $B$ become a linear identification of the forward dynamics. Finally, we obtain the \textit{analytical} solution $A$ and $B$ matrices by solving the optimal solution of $J(K)$ in Equation~\ref{equ:dynamics_loss}, which in general only takes seconds of CPU-only computation.
To promote stable long-horizon predictions, we apply \textit{spectral clipping}~\cite{guo2026surprisingeffectivenessspectralclipping} as a post-hoc step.
Any eigenvalue of $A$ whose magnitude exceeds one is clipped to one without altering the corresponding eigenvectors, resulting in a provably stable linear system.

Once we obtain the linear evolution in the lifting space, we can incorporate the existing control framework for high-level tasks. In our case, we leverage the existing linear Model-Predictive Control algorithm to design the safety-guaranteed legged navigation controller. The overview is illustrated in Fig.~\ref{fig:navigation}.
Our optimization can be formulated as follows:

\[
\begin{aligned}
\min_{\phi_{1 \cdots H}, \bm{u}_{1 \cdots H}} \quad & J(\phi_{1 \cdots H}, \bm{u}_{1 \cdots H}) \\
\text{s.t.} \quad & \phi_{k+1} = A \phi_k + B \bm{u}_k \quad \forall k = 0, \dots, H-1\\
& h_i(\phi_k) \geq 0 \quad \forall h_i \in \mathcal{H} \text{ and } \forall k = 1, \dots, H \\
& \bm{u}_k \in \mathcal{U}.
\end{aligned}
\]
The key idea of our framework is to directly solve lifted variables $\phi_{1 \cdots H}$, instead of the original states $\bm{x}_{1 \cdots H}$, along with control variables $\bm{u}_{1 \cdots H}$. This formulation is effective because all the dynamics become linear in the lifted space.

Our objective function is designed to track the given trajectory $\hat{\bm{x}}_{1 \cdots H}$ in the original state space:
\[
J(\phi_{1 \cdots H}, \bm{u}_{1 \cdots H}) = \sum_{k=1}^{H} ||\bm{x}_k - \hat{\bm{x}}_k||_P + \sum_{k=0}^{H-1} ||\bm{u}_{k+1} - \bm{u}_k||_2,
\]
where $\bm{x}_k = \phi^{-1}(\phi_k)$ for all timesteps $k$. Designing the lifting function $\phi$ to include the original state information, the inverse operation $\phi^{-1}(\cdot)$ becomes a linear projection, making the cost function quadratic with respect to all the variables.

The first equality constraint aims to satisfy the learned dynamics. The second inequality constraint is an arbitrary task-specific constraint, in our case, a distance barrier function $h_i(\phi_k) = \hat{h}_i(\bm{x}_k) = \textbf{dist}(\bm{x}_k, \text{obstacle}_i)$ for all obstacles. The third constraint confines control signals within the valid region.

Note that this formulation can readily incorporate safety frameworks such as discrete control barrier functions~\cite{zeng2021safety}, by modifying direct collision checking $\hat{h}(\bm{x}_k) \geq 0$ constraint to that of control barrier function: $\hat{h}(\bm{x}_{k+1}) \geq \gamma \hat{h}(\bm{x}_k) \quad$ for $\gamma \in [0, 1)$, which can still be linear based on obstacle geometry and choice of barrier function $\hat{h}$.

\section{Experimental Setup}
\label{sec:exp}

We validate our framework in both simulation and hardware. Our simulation experiments, including data collection and safe navigation evaluations, are conducted using the Unitree G1~\cite{unitree_g1} robot in the IsaacGym simulator~\cite{makoviychuk2021isaac}. 
The MPC optimization is solved using IPOPT interfaced through CasADi~\cite{Andersson2019}.

\subsection{Low-level Locomotion Policy}
Our high-level navigation framework computes control actions at the base-velocity level, making it agnostic to the underlying whole-body controller. This allows seamless deployment over different black-box locomotion policies without modification to the MPC formulation.

For all simulation experiments, we train a velocity-tracking locomotion controller using Proximal Policy Optimization (PPO) following the reward structure from~\cite{unitree_rl_gym}.
The low-level control policy $\pi_\theta(\bm{a}_t \mid \bm{o}_t)$ maps proprioceptive observations $\bm{o}_t$ to actions $\bm{a}_t \in \mathbb{R}^{12}$, which correspond to target joint angles for the 12 lower body joint actuators. A PD controller then converts these targets into torques using joint-specific stiffness and damping gains. The policy observes $\bm{o}_t \in \mathbb{R}^{47}$:
$$\bm{o}_t = [\bm{\omega}_t, \bm{g}_t, \hat{\bm{v}}, \bm{q},\dot{\bm{q}}_t, \bm{a}_{t-1}, \sin(2\pi c_t), \cos(2\pi c_t)]$$
The observation vector consists of the base angular velocity $\bm{\omega}_t \in \mathbb{R}^3$, the projected gravity vector $\bm{g}_t \in \mathbb{R}^3$, and the command linear and angular velocities $\hat{\bm{v}} = [\hat{v}_x, \hat{v}_y, \hat{\omega}] \in \mathbb{R}^3$. It further includes the joint positions and velocities $\bm{q}, \dot{\bm{q}}\in\mathbb{R}^{12}$, together with the previous action $\bm{a}_{t-1} \in \mathbb{R}^{12}$. To aid in learning rhythmic gaits, a 2D gait phase clock $[\sin(2\pi c_t), \cos(2\pi c_t)]$ is included, where the phase clock $c_t$ cycles with a period of 0.8~seconds. The policy is modeled using a recurrent neural network (LSTM).

Training is performed in the IsaacGym simulator~\cite{makoviychuk2021isaac} with 4,096 parallel environments for 7,000 steps, requiring approximately two hours on an NVIDIA RTX 4090 GPU.

\subsection{Locomotion Data Collection}

To learn the Koopman forward model, we collect a dataset of high-level state transitions. We sample velocity commands $\bm{u}$ every 0.5 seconds using three strategies following \cite{kim2022learning}: (1) \textit{uniform random}, (2) \textit{linearly correlated}, and (3) \textit{normally correlated}.
We collect 50 trajectories per strategy, yielding 150 trajectories in total, each comprising 150 seconds of robot state recorded at 50 Hz.

Since the MPC planning rollout begins from the robot's current local frame, we improve distribution alignment by extracting sliding windows of $H_{\text{seq}}=16$ steps from each trajectory and transforming all states into the local coordinate frame of the first state in the window.
From this augmentation, we obtain a final dataset of $236,250$ state transitions.

\subsection{Koopman Lifting Function Design}
We evaluate several classes of state-lifting functions to identify the most effective representation for bipedal locomotion dynamics.
Concretely, for state $\bm{x} \in \mathbb{R}^n$, the lifting function takes the form:
\begin{equation*}
\phi(\bm{x}) = \Bigl[\bm{x},\;
\underbrace{\bm{x}^2, \ldots, \bm{x}^d}_{\text{poly}},\;
\underbrace{x_i x_j}_{i < j}^{\text{cross}},\;
\underbrace{\sin(\bm{x}),\, \cos(\bm{x})}_{\text{trig}},\;
1\Bigr],
\end{equation*}
where $d$ is the maximum polynomial degree and all operations are applied element-wise.
The full observable is $\psi(\bm{x}_t, \bm{u}_t) = [\phi(\bm{x}_t);\, \bm{u}_t]$, consistent with Section~\ref{sec:preliminary}.
We search over polynomial degrees $d \in \{1, 2, 3\}$ and the inclusion of cross-product and trigonometric terms, selecting the best combination based on validation-set performance.

\begin{table*}[htp]
    \centering
    \newcommand{\best}[1]{\textbf{#1}}
    \setlength{\tabcolsep}{3pt}
    \resizebox{\linewidth}{!}{%
    \begin{tabular}{lccccccccccc}
        \toprule
        \textbf{Model}
            & $\mathbf{RMSE_{phys}}$
            & \textbf{Train}
            & \textbf{Lift}
            & \textbf{RMSE}
            & \textbf{RMSE}
            & \textbf{RMSE}
            & \textbf{RMSE}
            & \textbf{RMSE}
            & \textbf{RMSE}
            & \textbf{RMSE}
            & \textbf{RMSE} \\
        &
            & \textbf{Time (s)}
            & \textbf{Dim}
            & \textbf{Pos.\ (m)}
            & \textbf{Vel.\ (m/s)}
            & \textbf{X (m)}
            & \textbf{Y (m)}
            & \textbf{Yaw (rad)}
            & \textbf{Vx (m/s)}
            & \textbf{Vy (m/s)}
            & \textbf{Yaw Rate (rad/s)} \\
        \midrule
        \multicolumn{12}{l}{\textit{\textbf{Koopman Lifting Models}}} \\[1pt]
        \multicolumn{12}{l}{\textit{Phase Augmented (top 3 by $\text{RMSE}_\text{phys}$)}} \\
        \quad Koopman-PA (Poly3+Cross+Trig) & \best{0.1616} & 0.59         & 72          & 0.0950        & 0.0871        & 0.0928        & \best{0.0971} & \best{0.1348} & \best{0.0714} & 0.1003        & \best{0.3246} \\
        \quad Koopman-PA (Poly2+Cross+Trig) & 0.1621        & 0.55         & 64          & 0.0950        & \best{0.0869} & 0.0928        & 0.0972        & 0.1380        & 0.0715        & \best{0.1000} & 0.3247        \\
        \quad Koopman-PA (Poly3+Cross)       & 0.1622        & 0.43         & 56          & \best{0.0949} & 0.0871        & \best{0.0927} & 0.0972        & 0.1386        & 0.0716        & 0.1001        & 0.3247        \\
        \multicolumn{12}{l}{\textit{No Phase (top 3 by $\text{RMSE}_\text{phys}$)}} \\
        \quad Koopman-NP (Poly3+Cross+Trig) & 0.1697        & 0.34         & 49          & 0.0978        & 0.1123        & 0.0942        & 0.1012        & 0.1378        & 0.0729        & 0.1410        & 0.3309        \\
        \quad Koopman-NP (Poly2+Cross+Trig) & 0.1702        & 0.28         & 43          & 0.0978        & 0.1122        & 0.0942        & 0.1013        & 0.1409        & 0.0729        & 0.1409        & 0.3311        \\
        \quad Koopman-NP (Poly3+Cross)       & 0.1702        & \best{0.24}  & \best{37}   & 0.0978        & 0.1122        & 0.0941        & 0.1013        & 0.1414        & 0.0731        & 0.1408        & 0.3308        \\
        \midrule
        \multicolumn{12}{l}{\textit{\textbf{Baseline Models}}} \\
        \quad Linear-PA & \best{0.1762} & 0.016        & --- & \best{0.1439} & \best{0.0894} & \best{0.1404} & \best{0.1473} & \best{0.1429} & \best{0.0739} & \best{0.1026} & \best{0.3294} \\
        \quad Linear                   & 0.1821        & \best{0.007} & --- & 0.1449        & 0.1135        & \best{0.1404} & 0.1493        & 0.1436        & 0.0740        & 0.1424        & 0.3326        \\
        \quad Component-Wise Linear    & 0.1831        & 0.016        & --- & 0.1450        & 0.1136        & \best{0.1404} & 0.1493        & 0.1440        & 0.0743        & 0.1425        & 0.3354        \\
        \quad MLP-PA    & 0.1974        & 296.4        & --- & 0.1888        & 0.1207        & 0.2024        & 0.1741        & 0.1503        & 0.1075        & 0.1326        & 0.3329        \\
        \quad MLP                      & 0.1987        & 298.2        & --- & 0.1852        & 0.1310        & 0.2059        & 0.1619        & 0.1408        & 0.0986        & 0.1568        & 0.3378        \\
        \quad Integrator               & 0.2888        & ---          & --- & 0.1753        & 0.1725        & 0.1775        & 0.1731        & 0.2355        & 0.1537        & 0.1894        & 0.5693        \\
        \bottomrule
    \end{tabular}}
    \caption{Average \textit{next-step prediction error} on the validation dataset.
    Rows are sorted by $\text{RMSE}_\text{phys}$ ($\downarrow$) within each group.
    Best values in each group are highlighted in \textbf{bold}.
    }
    \label{tab:dynamics_model_comparison}
\vspace{-8pt}
\end{table*}

\noindent\textbf{Phase Augmentation.}
Bipedal locomotion is inherently periodic: the robot's base-level motion is tightly coupled to its internal gait phase.
The locomotion policy already encodes this periodicity through a gait phase clock $[\sin(2\pi c_t), \cos(2\pi c_t)]$ in its observation (Sec.~\ref{sec:exp}).
To expose this structure to the Koopman model, we augment the nominal 6-dimensional base state $\bm{x}_t$ with the phase clock, yielding an 8-dimensional input
$\tilde{\bm{x}}_t = [\bm{x}_t;\, \sin(2\pi c_t),\, \cos(2\pi c_t)] \in \mathbb{R}^8$,
which is then passed to $\phi(\cdot)$.
We refer to models trained with this augmentation as \textit{Phase Augmented} (PA), and those using only the base state as \textit{No Phase} (NP).

\begin{figure}[t]
\vspace{20pt}
\centering
     \begin{subfigure}[b]{0.49\columnwidth}
         \centering
         \includegraphics[width=\textwidth]{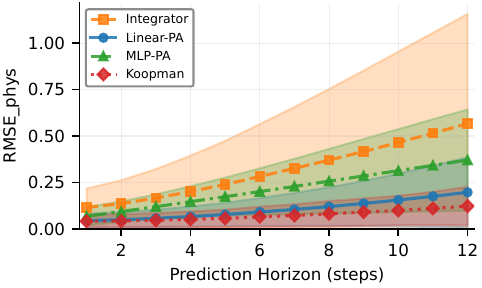}
         \caption{Long-term State error}
         \label{fig:rmse_phys}
     \end{subfigure}
     \hfill
     \begin{subfigure}[b]{0.49\columnwidth}
         \centering
         \includegraphics[width=\textwidth]{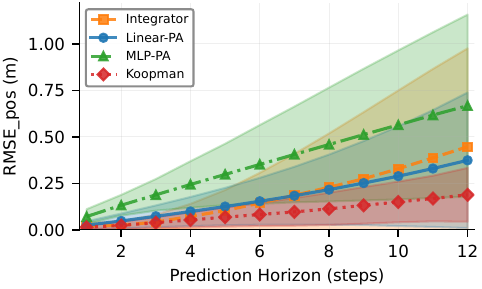}
         \caption{Long-term Position error}
         \label{fig:rmse_pos}
     \end{subfigure}
     \caption{Long-term prediction error comparison. Solid lines and shaded regions denote the mean and standard deviation, respectively.}
     \label{fig:longterm_prediction_batch}
\vspace{15pt}
\end{figure}

\subsection{Evaluation Baselines}
We compare the \textit{Koopman} model against the following dynamics baselines:

\noindent\textit{Integrator}: An ideal first-order model: 
$[p_x, p_y, \theta]_{t+1} = [p_x, p_y, \theta]_{t} + dt [v_x, v_y, \omega]_{t}$; $[v_x, v_y, \omega]_{t+1} = [\hat{v}_x, \hat{v}_y, \hat{\omega}]_{t}$, 
where $dt$ represents the duration of one simulation step, and velocities are accumulated in the global frame. Note that this baseline does not require data for training.

\noindent\textit{Component-wise Linear (Comp.Lin.)}: Building on the baseline in~\cite{li2022bridging}, we learn a block-structured linear model by assuming independence among the components in the state dynamics.

\noindent\textit{Linear}: This baseline assumes $\phi(\bm{x}) = \bm{x}$ (i.e., no other lifting functions), yielding a linear system in the original state space.
We evaluate two variants: \textit{Linear-NP} (No Phase, trained on the 6D base state) and \textit{Linear-PA} (Phase Augmented, trained on the 8D state including the gait clock), following the same phase-augmentation scheme as the Koopman models (Sec.~\ref{sec:exp}).

\noindent\textit{Multi-layer Perceptron (MLP)}: We train an MLP for 400 epochs to learn the input and output of the transition function, as described in Sec.~\ref{subsec:learning_dynamics}. Our network has two hidden layers of [32, 32] with a ReLU activation function.
As with the linear model, we evaluate both \textit{MLP-NP} and \textit{MLP-PA} variants.

\begin{figure}[t]
    \centering
    \hspace*{-1cm} 
    \includegraphics[width=0.35\textwidth]{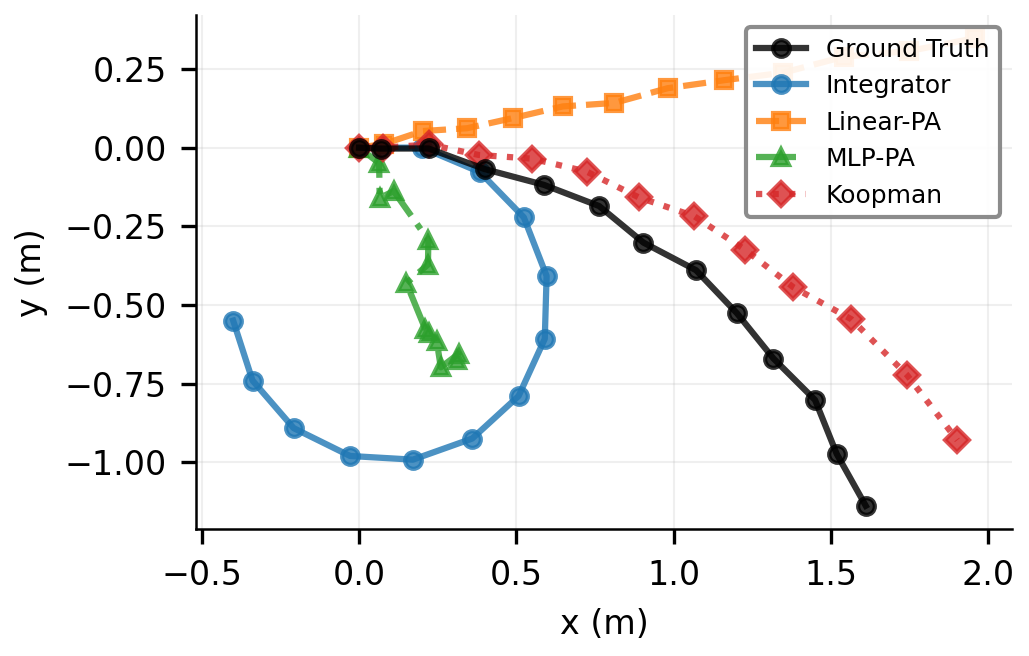}
\vspace{-8pt}
    \caption{Long-term trajectory predictions of Koopman model against baselines. The ground-truth trajectory is collected under a constant command held out from training.
    }
    \label{fig:longterm_prediction_single}
\end{figure}

\section{Results}
\label{sec:results}

\subsection{Does Phase-Aware Koopman Lifting Provide the Best One-Step Predictive Accuracy?}\label{subsec:prediction}

We train various Koopman dynamics models and compare them against several baselines, as summarized in Table~\ref{tab:dynamics_model_comparison}.
The evaluation metrics include RMSE$_\text{phys}$, training time, and lifted dimension, where RMSE$_\text{phys}$ is computed over the physical state dimensions only (excluding the phase augmentation).
Overall, these results demonstrate that Koopman lifting consistently outperforms all baseline approaches in predictive accuracy.
The best model, \emph{Koopman-PA (Poly3+Cross+Trig)}, achieves an RMSE$_\text{phys}$ of $0.1616$, outperforming the best linear baseline (\emph{Linear-PA}: $0.1762$) and the best MLP baseline (\emph{MLP-PA}: $0.1974$).
The most significant gain is in positional error, which is reduced by $34\%$ relative to \emph{Linear-PA}.
Furthermore, incorporating phase augmentation notably improves predictive performance; compared to the No Phase (NP) formulation, the Phase-Augmented (PA) Koopman model reduces the 1-step physical state error and velocity error by approximately 5\% and 23\%, respectively.

For the remainder of the experiments, we adopt \emph{Koopman-PA (Poly3+Cross+Trig)} as our proposed model, which combines phase augmentation, degree-3 polynomial, cross-product, and trigonometric lifting, strictly adhering to the formulation $\psi(\bm{x}_t, \bm{u}_t) = [\phi(\bm{x}_t),\, \bm{u}_t]$ in Section~\ref{sec:preliminary}.

\subsection{Does the Koopman Model Maintain Accuracy Over Long-Horizon Rollouts?}
Prediction errors accumulate rapidly in long-horizon rollouts.
In Fig.~\ref{fig:longterm_prediction_batch}, we show the error on validation trajectories with uniform random commands sampled every 0.5 seconds.
After a 12-step rollout (6 seconds), \emph{Koopman} achieves an RMSE$_\text{phys}$ of $0.123$, compared to $0.197$ for \emph{Linear-PA} and $0.370$ for \emph{MLP-PA}.
The advantage is further amplified in positional error: the RMSE$_\text{pos}$ of \emph{Koopman-PA} after 12 steps is $0.188$\,m, a $50\%$ reduction compared to \emph{Linear-PA} ($0.374$\,m) and a $72\%$ reduction compared to \emph{MLP-PA} ($0.668$\,m).

Fig.~\ref{fig:longterm_prediction_single} illustrates a held-out constant-command trajectory, comparing rollout predictions from \emph{Koopman-PA}, \emph{Linear-PA}, \emph{MLP-PA}, and \emph{Integrator}.
The \emph{Integrator} reacts instantaneously to velocity commands and therefore overshoots sharp turns, while \emph{Linear-PA} undershoots them due to its limited model capacity.
\emph{MLP-PA} diverges rapidly from the ground truth after the first few steps.
By contrast, \emph{Koopman-PA} closely tracks the ground-truth trajectory throughout the entire rollout.

\begin{figure}[t]
    \centering
    \begin{subfigure}[b]{0.45\textwidth}
        \centering
        \includegraphics[width=\textwidth]{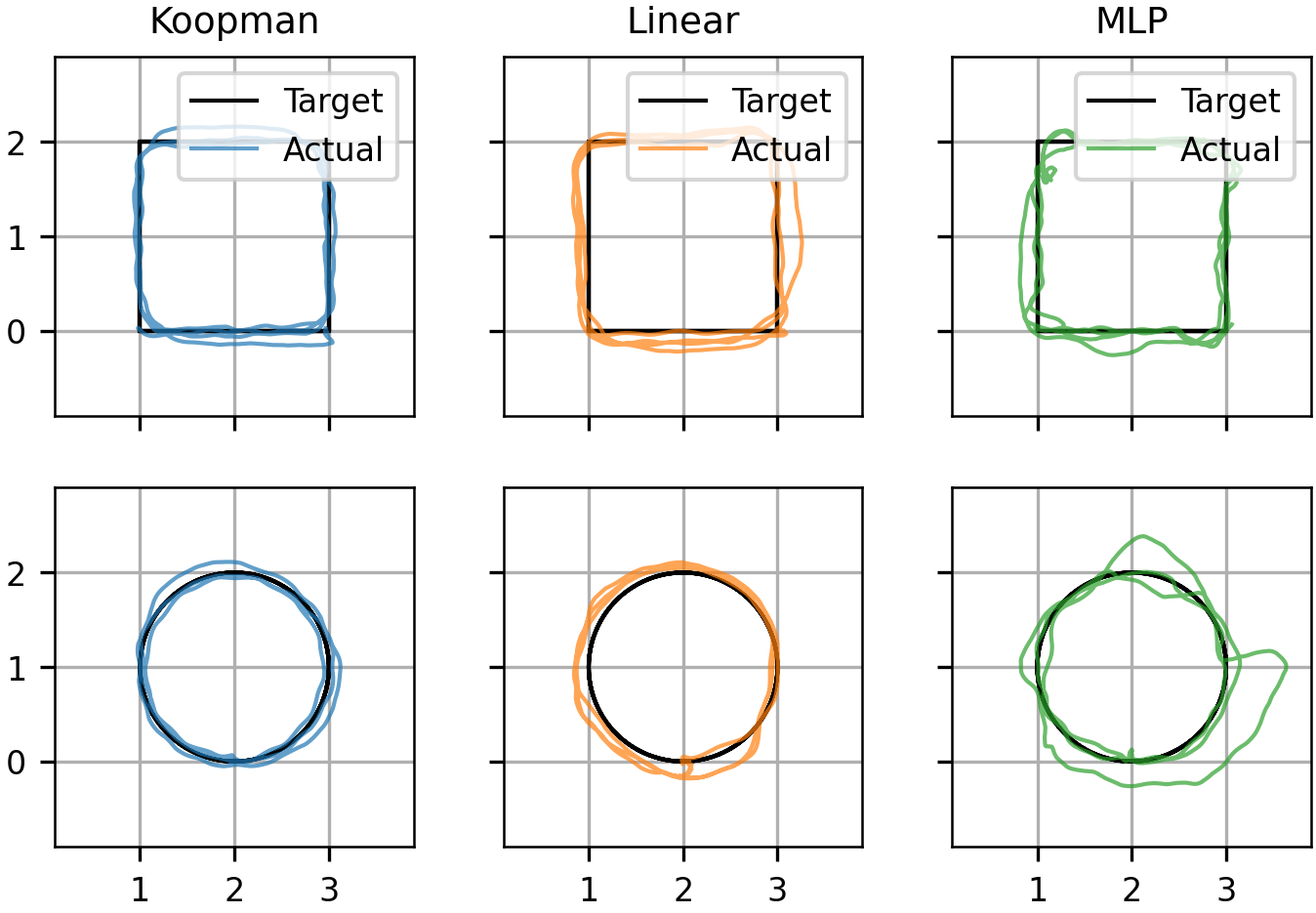}
        \caption{Path-tracking trajectories.}
        \label{fig:opentracking_traj}
    \end{subfigure}

    \begin{subfigure}[b]{0.22\textwidth}
        \centering
        \includegraphics[width=\textwidth]{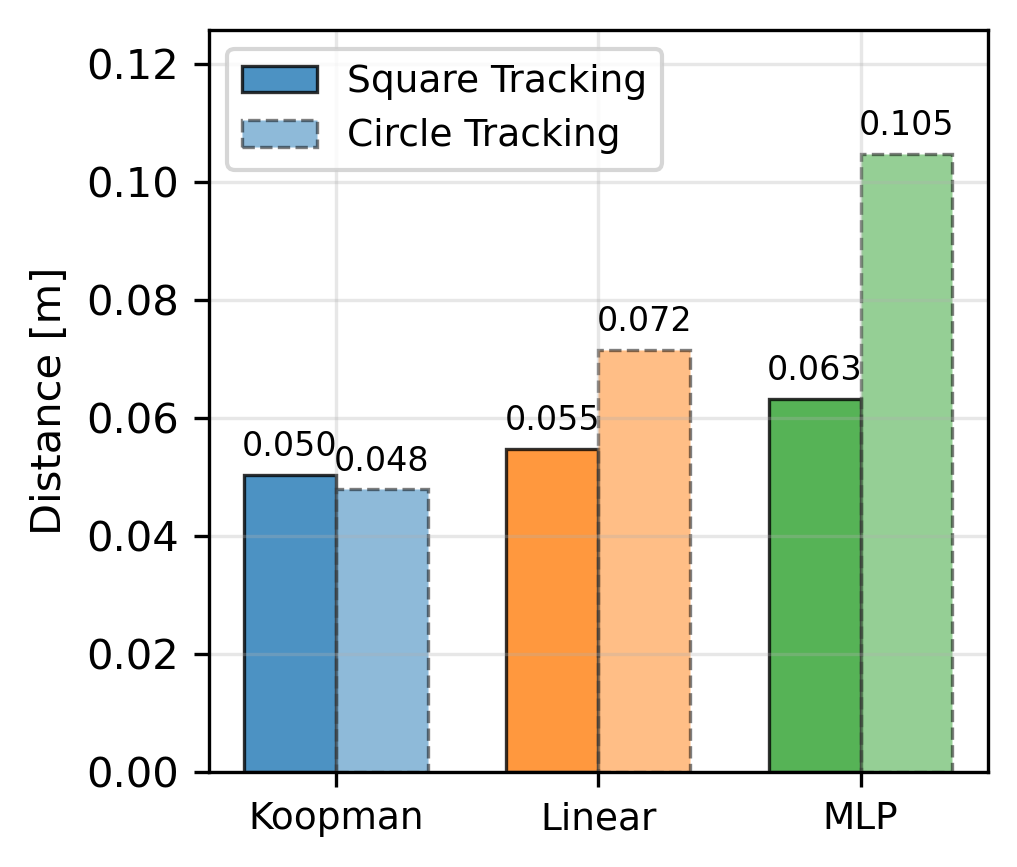}
        \caption{Avg.\ distance from path.}
        \label{fig:opentracking_dist}
    \end{subfigure}
    \begin{subfigure}[b]{0.22\textwidth}
        \centering
        \includegraphics[width=\textwidth]{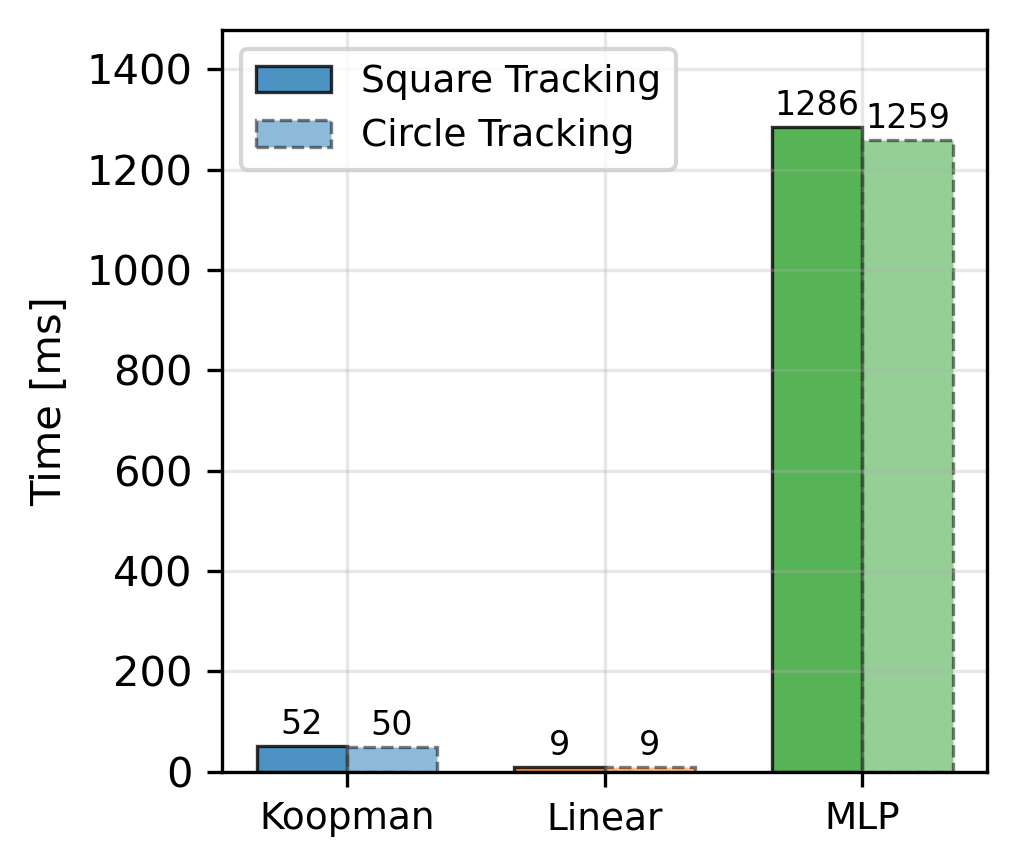}
        \caption{MPC solver time.}
        \label{fig:opentracking_solver}
    \end{subfigure}
    \caption{Open Space Path Tracking Comparison.}
    \label{fig:opentracking}
\end{figure}

\subsection{Does Koopman MPC Achieve More Accurate Path Tracking Than The Baselines?}
\label{subsec:planning}


We compare tracking performance and required computation time of \emph{Koopman} model against two baselines without phase augmentation: \emph{Linear} and \emph{MLP}.
The three models are evaluated on tracking two trajectories, a square and a circle, both involving multiple turns.
The task ends when the robot loops 3 times.
The result is shown in Fig.~\ref{fig:opentracking}.
Fig.~\ref{fig:opentracking}(a) shows the reference and actual trajectories, Fig.~\ref{fig:opentracking}(b) reports the average distance from the reference path, and Fig.~\ref{fig:opentracking}(c) shows the MPC solver time per step.

From the MPC Solver time graph, we can see that our NLP solver~\cite{Andersson2019} with MLP as the dynamics equality constraint makes the problem infeasible for navigation tasks. Even for this simple Open Trajectory task and a relatively small network size, this leads to average solver time greater than 1.2 seconds, requiring more than 25 times larger solver time of our Koopman model with 72 lifted dimensions. This is due to the high computational cost of evaluating the neural network’s gradients and Hessians, as well as the non-convexity introduced by the MLP’s activation functions.

In terms of tracking performance, we observe that our \emph{Koopman} model achieves the best overall performance, with its trajectories overlapping with the reference paths. This aligns with the higher accuracy of our model.
The MLP-based model performs the worst, producing unstable and erratic trajectories. This is largely because the nonlinear equality constraints make it difficult for the solver to find a stable solution and end up giving intermediate, not-yet fully optimized trajectories.
Quantitatively, our model achieves the lowest tracking errors of $0.050$ m and $0.048$ m in the square and circle tracking tasks, reducing errors by $13.8\%$ and $35.7\%$ compared to the second-best model, \emph{Linear-NP}.

\begin{table*}[t]
\centering
\resizebox{\textwidth}{!}{%
\begin{tabular}{l|ccc|ccc|ccc|ccc}
\toprule
 & \multicolumn{3}{c|}{\textbf{Integrator}}
 & \multicolumn{3}{c|}{\textbf{Linear (No Phase)}}
 & \multicolumn{3}{c|}{\textbf{Linear (Phase Augmented)}}
 & \multicolumn{3}{c}{\textbf{Koopman (Ours)}} \\
\cmidrule(lr){2-4}\cmidrule(lr){5-7}\cmidrule(lr){8-10}\cmidrule(lr){11-13}
 & SR (\%) & $\bar{p}$ (m) & $p_{\max}$ (m)
 & SR (\%) & $\bar{p}$ (m) & $p_{\max}$ (m)
 & SR (\%) & $\bar{p}$ (m) & $p_{\max}$ (m)
 & SR (\%) & $\bar{p}$ (m) & $p_{\max}$ (m) \\
\midrule
Corridor 1 ($n{=}20$) & 55  & 0.0127 & 0.0684 & 95  & 0.0042 & 0.0480 & \textbf{100} & 0.0038 & 0.0414 & 95  & \textbf{0.0013} & \textbf{0.0146} \\
Corridor 2 ($n{=}20$) & 60  & 0.0134 & 0.0608 & \textbf{100} & 0.0069 & 0.0389 & 95  & 0.0035 & 0.0322 & 95  & \textbf{0.0018} & \textbf{0.0199} \\
Maze 1     ($n{=}5$)  & 80  & 0.0089 & 0.1125 & 40  & 0.0087 & 0.0879 & 20  & \textbf{0.0065} & \textbf{0.0643} & \textbf{100}  & 0.0132 & 0.0756 \\
Maze 2     ($n{=}5$)  & 60  & 0.0223 & 0.1650 & 20  & \textbf{0.0116} & 0.1102 & 60  & 0.0179 & 0.0903 & \textbf{100}  & 0.0172 & \textbf{0.0674} \\
\midrule
Total      ($n{=}50$) & 60  & 0.0136 & 0.0794 & 84  & 0.0065 & 0.0546 & 86  & 0.0054 & 0.0449 & \textbf{96}  & \textbf{0.0043} & \textbf{0.0281} \\
\bottomrule
\end{tabular}%
}
\caption{Safe navigation results across four environments and in total.
SR = Success Rate ($\uparrow$); $\bar{p}$ = mean geometrical violation depth ($\downarrow$); $p_{\max}$ = maximum violation depth ($\downarrow$).
Best value per row per metric is \textbf{bold}.}
\label{tab:navigation}
\vspace{-15pt}
\end{table*}

\subsection{Does the Better Prediction of the Koopman Model Translate to Safer Navigation in Cluttered Environments?}
A core advantage of our safe navigation framework is its easy extensionability to unseen environments.
In this section, we compare the safe navigation performance of MPC with the learned \emph{Koopman} dynamics against the \emph{Integrator} and \emph{Linear} models. The \emph{MLP} model is omitted due to its poor tracking performance and high computational cost.
We evaluate our framework in four safe navigation environments:

\begin{figure}
    \centering
    \includegraphics[width=0.48\textwidth]{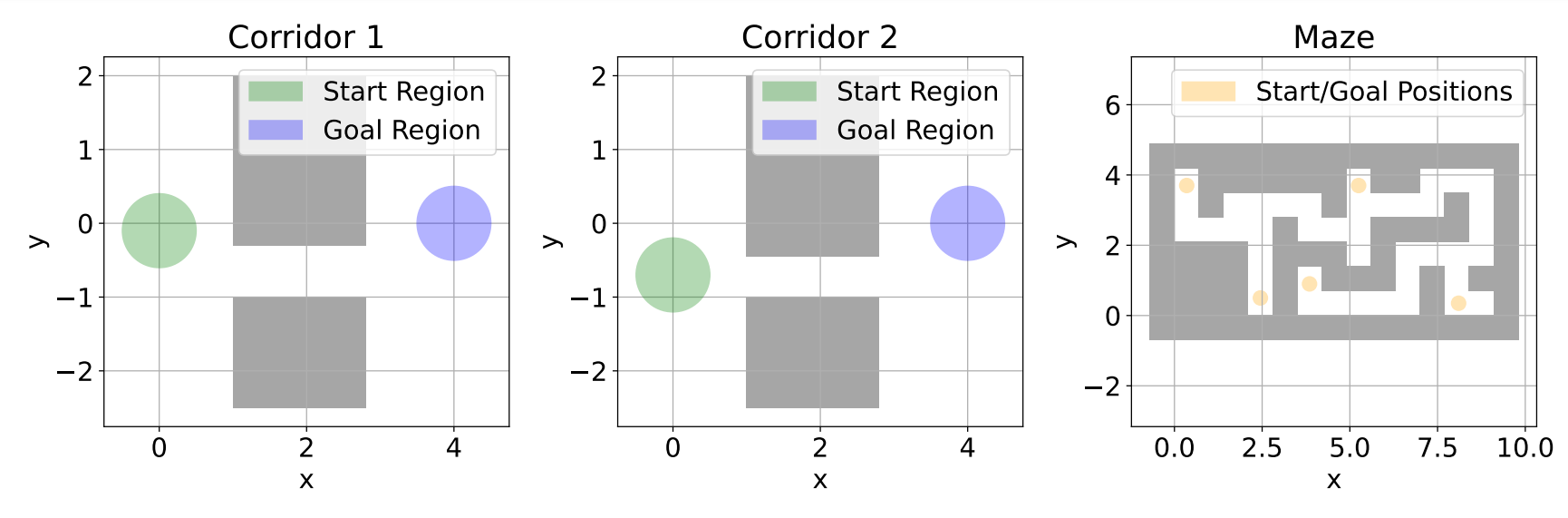}
    \caption{Environments used for Navigation. For \textit{corridor 1} and \textit{corridor 2}. Green and purple indicate the start and goal regions. For \textit{maze} environment, start and goal positions are sampled from orange points.}
    \label{fig:maps}
\end{figure}


\noindent\textbf{i)} \textit{Corridor 1}, a corridor with a 70 cm gap

\noindent\textbf{ii)} \textit{Corridor 2}, a narrower corridor with a 55 cm gap

\noindent\textbf{iii)} \textit{Maze 1}, an artifact maze~\cite{thirugnanam2022safety} with 90 cm passages

\noindent\textbf{iv)} \textit{Maze 2}, the same maze with narrower passages of 80 cm.

For the \textit{Corridor} environments, we randomly sample 20 pairs of start and goal positions across the corridor.
The goal is to reach the goal by crossing the corridor in 10 seconds.
For the \textit{Maze} environments, we sample 5 start and goal position pairs, and the task is to reach the goal within 50 seconds.
This totals up to 50 navigation scenarios.

To isolate planning performance from perception, we assume obstacle positions and shapes are known. A high-level planner generates sparse waypoints using A*~\cite{hart1968formal}, which serve as reference trajectories for the MPC controller.

We report the Success Rate (\%) and average, maximum violation(m) in Table~\ref{tab:navigation}.
Success Rate (\%) measures whether the robot was able to reach the goal without getting stuck.
This happens when the MPC planner is not able to plan a collision-free path towards the goal.
The average and maximum violation depth is measured based on the robot's safety range specified by a 40cm circle for \textit{Corridor} tasks, and two 30cm circles for the \textit{Mazes} tasks.

The \emph{Koopman} model was able to succeed in $96\%$ of the environments.
In particular, it reaches success rates of 100\% in \emph{Maze} environments.
This significantly outperforms the average of 30\% and 40\% for linear models, and 70\% for the Integrator model.
This improvement arises because \emph{Koopman} models are better at capturing turning commands compared to baselines, as \emph{Maze} environment requires frequent turning.
\emph{Linear} models are not good at capturing turning motions, and thus result in a lower success rate compared to the \emph{Integrator} model despite their better prediction accuracy.

\begin{figure}[t]
    \centering
    \includegraphics[width=0.48\textwidth]{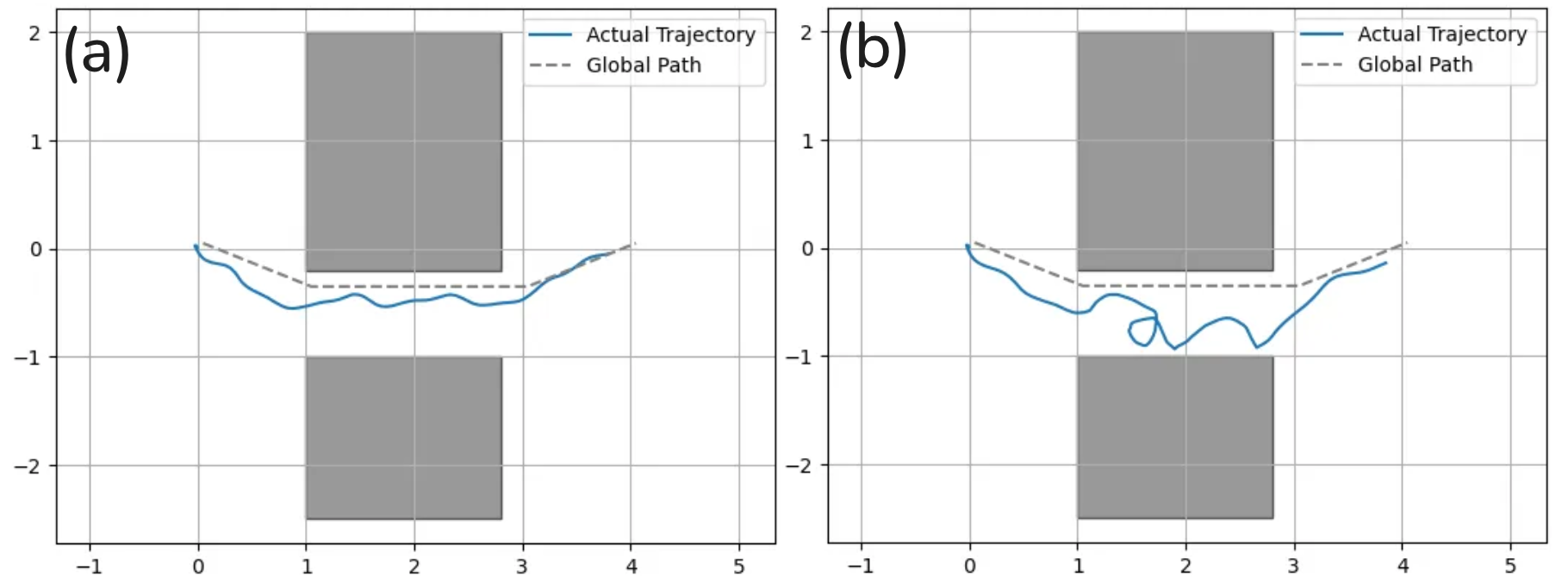}
    \caption{
    Navigation trajectory in \emph{Corridor 1}. (a) The \emph{Koopman} model successfully passes through the corridor without collision, while (b) the \emph{Linear} baseline collides with the walls. 
    }
    \label{fig:safe_nav_traj}
\end{figure}

For the \emph{Corridor} environments, both \emph{Linear} and \emph{Koopman} models achieve near-perfect success rate.
However, the high violation depths imply that the robot was not able to maintain safe distance during its path as can be seen in Figure~\ref{fig:safe_nav_traj}.

\subsection{Can Koopman MPC successfully perform safe navigation on hardware?}
To validate the feasibility of the proposed framework on a physical system, we deploy our pipeline on the Unitree G1 hardware robot.
As our framework is agnostic to the underlying locomotion policy, we use the pretrained locomotion policy from \cite{ben2025homie} for the hardware experiment.
We first collect approximately 10 minutes of real-world locomotion data by commanding the robot to move freely in an empty lab space.
This data is used to fit the Koopman forward model following the same procedure described in Sec.~\ref{subsec:learning_dynamics}.
For the MPC, we use a planning horizon of $H = 4$ steps at $\Delta t = 0.5\,\text{s}$ per step, yielding a 2-second look-ahead.
Fig.~\ref{fig:safe_navigation}(c) shows the robot during a representative validation run.

\label{sec:summary}
\section{Conclusion and Future Work}
In this work, we present a safe navigation framework for bipedal robots by combining Koopman operator theory with the MPC framework. We first train a low-level locomotion control policy using deep reinforcement learning. Then, we learn the nonlinear base-level dynamics from the collected trajectories using Dynamic Mode Decomposition. Finally, we develop an MPC framework for safe navigation that efficiently solves the given constrained optimization problem by leveraging the linearity of the learned Koopman dynamics. We show that the learned Koopman dynamics offer improved accuracy for long-term forward dynamics prediction.
We further deploy the framework in narrow-passage corridor and maze environments, showcasing its effectiveness across diverse challenging scenarios.

There are several directions for future work. For instance, the utility of the learned linear dynamics is not limited to obstacle avoidance discussed in this work. There exists a wide range of potential tasks, such as loco-manipulation, or traversing more complex environments, which require modeling full-body dynamics. This goal will likely require a careful investigation of the lifting function for the reliable Koopman linearization. We also aim to extend the capability of this method by exploring diverse lifting techniques. One possibility is to utilize time-delay or recurrent neural network based lifting, which could provide a more concise and accurate forward dynamics model. However, this may introduce additional complexity into the learning framework. Investigating this trade-off will be an interesting future research direction for legged robots.

\bibliographystyle{IEEEtran}  
\bibliography{references}  

@article{jung2025ppf,
  title={{PPF}: Pre-training and preservative fine-tuning of humanoid locomotion via model-assumption-based regularization},
  author={Jung, Hyunyoung and Gu, Zhaoyuan and Zhao, Ye and Park, Hae-Won and Ha, Sehoon},
  journal={IEEE Robotics and Automation Letters},
  year={2025},
  publisher={IEEE}
}

@inproceedings{xie2022glide,
  title={{GLIDE}: Generalizable quadrupedal locomotion in diverse environments with a centroidal model},
  author={Xie, Zhaoming and Da, Xingye and Babich, Buck and Garg, Animesh and de Panne, Michiel van},
  booktitle={International workshop on the algorithmic foundations of robotics},
  pages={523--539},
  year={2022},
  organization={Springer}
}

@article{makoviychuk2021isaac,
  title={{Isaac Gym}: High-performance {GPU}-based physics simulation for robot learning},
  author={Makoviychuk, Viktor and Wawrzyniak, Lukasz and Guo, Yunrong and Lu, Michelle and Storey, Kier and Macklin, Miles and Hoeller, David and Rudin, Nikita and Allshire, Arthur and Handa, Ankur and others},
  journal={arXiv preprint arXiv:2108.10470},
  year={2021}
}

@article{li2022bridging,
  title={Bridging model-based safety and model-free reinforcement learning through system identification of low dimensional linear models},
  author={Li, Zhongyu and Zeng, Jun and Thirugnanam, Akshay and Sreenath, Koushil},
  journal={arXiv preprint arXiv:2205.05787},
  year={2022}
}

@article{kim2022learning,
  title={Learning forward dynamics model and informed trajectory sampler for safe quadruped navigation},
  author={Kim, Yunho and Kim, Chanyoung and Hwangbo, Jemin},
  journal={arXiv preprint arXiv:2204.08647},
  year={2022}
}

@inproceedings{liao2023walking,
  title={Walking in narrow spaces: Safety-critical locomotion control for quadrupedal robots with duality-based optimization},
  author={Liao, Qiayuan and Li, Zhongyu and Thirugnanam, Akshay and Zeng, Jun and Sreenath, Koushil},
  booktitle={2023 IEEE/RSJ International Conference on Intelligent Robots and Systems (IROS)},
  pages={2723--2730},
  year={2023},
  organization={IEEE}
}

@article{Koopman1931Koopman,
author = {B. O. Koopman },
title = {Hamiltonian systems and transformation in {Hilbert} space},
journal = {Proceedings of the National Academy of Sciences},
volume = {17},
number = {5},
pages = {315-318},
year = {1931},
doi = {10.1073/pnas.17.5.315}}

@article{schmid2010dynamic,
  title={Dynamic mode decomposition of numerical and experimental data},
  author={Schmid, Peter J},
  journal={Journal of fluid mechanics},
  volume={656},
  pages={5--28},
  year={2010},
  publisher={Cambridge University Press}
}

@article{hart1968formal,
  title={A formal basis for the heuristic determination of minimum cost paths},
  author={Hart, Peter E and Nilsson, Nils J and Raphael, Bertram},
  journal={IEEE transactions on Systems Science and Cybernetics},
  volume={4},
  number={2},
  pages={100--107},
  year={1968},
  publisher={IEEE}
}

@article{williams2015data,
  title={A data-driven approximation of the {Koopman} operator: Extending dynamic mode decomposition},
  author={Williams, Matthew O and Kevrekidis, Ioannis G and Rowley, Clarence W},
  journal={Journal of Nonlinear Science},
  volume={25},
  number={6},
  pages={1307--1346},
  year={2015},
  publisher={Springer}
}

@inproceedings{cheng2024extreme,
  title={Extreme parkour with legged robots},
  author={Cheng, Xuxin and Shi, Kexin and Agarwal, Ananye and Pathak, Deepak},
  booktitle={2024 IEEE International Conference on Robotics and Automation (ICRA)},
  pages={11443--11450},
  year={2024},
  organization={IEEE}
}

@article{zhuang2023robot,
  title={Robot parkour learning},
  author={Zhuang, Ziwen and Fu, Zipeng and Wang, Jianren and Atkeson, Christopher and Schwertfeger, Soeren and Finn, Chelsea and Zhao, Hang},
  journal={arXiv preprint arXiv:2309.05665},
  year={2023}
}

@inproceedings{kim2023armp,
  title={{ARMP}: Autoregressive motion planning for quadruped locomotion and navigation in complex indoor environments},
  author={Kim, Jeonghwan and Li, Tianyu and Ha, Sehoon},
  booktitle={2023 IEEE/RSJ International Conference on Intelligent Robots and Systems (IROS)},
  pages={2731--2737},
  year={2023},
  organization={IEEE}
}

@article{winkler2018gait,
    title={Gait and trajectory optimization for legged systems through phase-based end-effector parameterization},
    author={Winkler, Alexander W and Bellicoso, C Dario and Hutter, Marco and Buchli, Jonas},
    journal={IEEE Robotics and Automation Letters},
    volume={3},
    number={3},
    pages={1560--1567},
    year={2018},
    publisher={IEEE}
}

@inproceedings{wellhausen2021rough,
  title={Rough terrain navigation for legged robots using reachability planning and template learning},
  author={Wellhausen, Lorenz and Hutter, Marco},
  booktitle={2021 IEEE/RSJ International Conference on Intelligent Robots and Systems (IROS)},
  pages={6914--6921},
  year={2021},
  organization={IEEE}
}

@article{tonneau2018efficient,
  title={An efficient acyclic contact planner for multiped robots},
  author={Tonneau, Steve and Del Prete, Andrea and Pettr{\'e}, Julien and Park, Chonhyon and Manocha, Dinesh and Mansard, Nicolas},
  journal={IEEE Transactions on Robotics},
  volume={34},
  number={3},
  pages={586--601},
  year={2018},
  publisher={IEEE}
}

@inproceedings{di2018dynamic,
  title={Dynamic locomotion in the {MIT} {Cheetah} 3 through convex model-predictive control},
  author={Di Carlo, Jared and Wensing, Patrick M and Katz, Benjamin and Bledt, Gerardo and Kim, Sangbae},
  booktitle={2018 IEEE/RSJ international conference on intelligent robots and systems (IROS)},
  pages={1--9},
  year={2018},
  organization={IEEE}
}

@article{haarnoja2018soft,
  title={Soft actor-critic algorithms and applications},
  author={Haarnoja, Tuomas and Zhou, Aurick and Hartikainen, Kristian and Tucker, George and Ha, Sehoon and Tan, Jie and Kumar, Vikash and Zhu, Henry and Gupta, Abhishek and Abbeel, Pieter and others},
  journal={arXiv preprint arXiv:1812.05905},
  year={2018}
}

@inproceedings{rudin2022learning,
  title={Learning to walk in minutes using massively parallel deep reinforcement learning},
  author={Rudin, Nikita and Hoeller, David and Reist, Philipp and Hutter, Marco},
  booktitle={Conference on Robot Learning},
  pages={91--100},
  year={2022},
  organization={PMLR}
}

@inproceedings{yang2023neural,
  title={Neural volumetric memory for visual locomotion control},
  author={Yang, Ruihan and Yang, Ge and Wang, Xiaolong},
  booktitle={Proceedings of the IEEE/CVF Conference on Computer Vision and Pattern Recognition},
  pages={1430--1440},
  year={2023}
}

@article{hoeller2024anymal,
  title={{ANYmal} parkour: Learning agile navigation for quadrupedal robots},
  author={Hoeller, David and Rudin, Nikita and Sako, Dhionis and Hutter, Marco},
  journal={Science Robotics},
  volume={9},
  number={88},
  pages={eadi7566},
  year={2024},
  publisher={American Association for the Advancement of Science}
}

@article{hoeller2021learning,
  title={Learning a state representation and navigation in cluttered and dynamic environments},
  author={Hoeller, David and Wellhausen, Lorenz and Farshidian, Farbod and Hutter, Marco},
  journal={IEEE Robotics and Automation Letters},
  volume={6},
  number={3},
  pages={5081--5088},
  year={2021},
  publisher={IEEE}
}

@article{chen2024korol,
  title={{KOROL}: Learning visualizable object feature with {Koopman} operator rollout for manipulation},
  author={Chen, Hongyi and Abuduweili, Abulikemu and Agrawal, Aviral and Han, Yunhai and Ravichandar, Harish and Liu, Changliu and Ichnowski, Jeffrey},
  journal={arXiv preprint arXiv:2407.00548},
  year={2024}
}

@inproceedings{asselmeier2024hierarchical,
  title={Hierarchical experience-informed navigation for multi-modal quadrupedal rebar grid traversal},
  author={Asselmeier, Max and Ivanova, Jane and Zhou, Ziyi and Vela, Patricio A and Zhao, Ye},
  booktitle={2024 IEEE International Conference on Robotics and Automation (ICRA)},
  pages={8065--8072},
  year={2024},
  organization={IEEE}
}

@article{merel2018hierarchical,
  title={Hierarchical visuomotor control of humanoids},
  author={Merel, Josh and Ahuja, Arun and Pham, Vu and Tunyasuvunakool, Saran and Liu, Siqi and Tirumala, Dhruva and Heess, Nicolas and Wayne, Greg},
  journal={arXiv preprint arXiv:1811.09656},
  year={2018}
}

@article{glanois2024survey,
  title={A survey on interpretable reinforcement learning},
  author={Glanois, Claire and Weng, Paul and Zimmer, Matthieu and Li, Dong and Yang, Tianpei and Hao, Jianye and Liu, Wulong},
  journal={Machine Learning},
  pages={1--44},
  year={2024},
  publisher={Springer}
}

@article{he2024agile,
  title={Agile but safe: Learning collision-free high-speed legged locomotion},
  author={He, Tairan and Zhang, Chong and Xiao, Wenli and He, Guanqi and Liu, Changliu and Shi, Guanya},
  journal={arXiv preprint arXiv:2401.17583},
  year={2024}
}

@inproceedings{ames2019control,
  title={Control barrier functions: Theory and applications},
  author={Ames, Aaron D and Coogan, Samuel and Egerstedt, Magnus and Notomista, Gennaro and Sreenath, Koushil and Tabuada, Paulo},
  booktitle={2019 18th European control conference (ECC)},
  pages={3420--3431},
  year={2019},
  organization={IEEE}
}

@inproceedings{thirugnanam2022safety,
  title={Safety-critical control and planning for obstacle avoidance between polytopes with control barrier functions},
  author={Thirugnanam, Akshay and Zeng, Jun and Sreenath, Koushil},
  booktitle={2022 International Conference on Robotics and Automation (ICRA)},
  pages={286--292},
  year={2022},
  organization={IEEE}
}

@inproceedings{grandia2021multi,
  title={Multi-layered safety for legged robots via control barrier functions and model predictive control},
  author={Grandia, Ruben and Taylor, Andrew J and Ames, Aaron D and Hutter, Marco},
  booktitle={2021 IEEE International Conference on Robotics and Automation (ICRA)},
  pages={8352--8358},
  year={2021},
  organization={IEEE}
}

@article{choi2020reinforcement,
  title={Reinforcement learning for safety-critical control under model uncertainty, using control {Lyapunov} functions and control barrier functions},
  author={Choi, Jason and Castaneda, Fernando and Tomlin, Claire J and Sreenath, Koushil},
  journal={arXiv preprint arXiv:2004.07584},
  year={2020}
}

@inproceedings{han2023utility,
  title={On the utility of {Koopman} operator theory in learning dexterous manipulation skills},
  author={Han, Yunhai and Xie, Mandy and Zhao, Ye and Ravichandar, Harish},
  booktitle={Conference on Robot Learning},
  pages={106--126},
  year={2023},
  organization={PMLR}
}

@article{koopman1932dynamical,
  title={Dynamical systems of continuous spectra},
  author={Koopman, Bernard O and Neumann, J v},
  journal={Proceedings of the National Academy of Sciences},
  volume={18},
  number={3},
  pages={255--263},
  year={1932},
  publisher={National Acad Sciences}
}

@article{han2024learning,
  title={Learning prehensile dexterity by imitating and emulating state-only observations},
  author={Han, Yunhai and Chen, Zhenyang and Williams, Kyle A and Ravichandar, Harish},
  journal={IEEE Robotics and Automation Letters},
  year={2024},
  publisher={IEEE}
}

@article{abraham2017model,
  title={Model-based control using {Koopman} operators},
  author={Abraham, Ian and De La Torre, Gerardo and Murphey, Todd D},
  journal={arXiv preprint arXiv:1709.01568},
  year={2017}
}

@inproceedings{folkestad2022koopnet,
  title={{KoopNet}: Joint learning of {Koopman} bilinear models and function dictionaries with application to quadrotor trajectory tracking},
  author={Folkestad, Carl and Wei, Skylar X and Burdick, Joel W},
  booktitle={2022 International Conference on Robotics and Automation (ICRA)},
  pages={1344--1350},
  year={2022},
  organization={IEEE}
}

@inproceedings{folkestad2021koopman,
  title={{Koopman} {NMPC}: {Koopman}-based learning and nonlinear model predictive control of control-affine systems},
  author={Folkestad, Carl and Burdick, Joel W},
  booktitle={2021 IEEE International Conference on Robotics and Automation (ICRA)},
  pages={7350--7356},
  year={2021},
  organization={IEEE}
}

@article{folkestad2020data,
  title={Data-driven safety-critical control: Synthesizing control barrier functions with {Koopman} operators},
  author={Folkestad, Carl and Chen, Yuxiao and Ames, Aaron D and Burdick, Joel W},
  journal={IEEE Control Systems Letters},
  volume={5},
  number={6},
  pages={2012--2017},
  year={2020},
  publisher={IEEE}
}

@article{bock1984multiple,
  title={A multiple shooting algorithm for direct solution of optimal control problems},
  author={Bock, Hans Georg and Plitt, Karl-Josef},
  journal={IFAC Proceedings Volumes},
  volume={17},
  number={2},
  pages={1603--1608},
  year={1984},
  publisher={Elsevier}
}

@inproceedings{giftthaler2018family,
  title={A family of iterative {Gauss-Newton} shooting methods for nonlinear optimal control},
  author={Giftthaler, Markus and Neunert, Michael and St{\"a}uble, Markus and Buchli, Jonas and Diehl, Moritz},
  booktitle={2018 IEEE/RSJ International Conference on Intelligent Robots and Systems (IROS)},
  pages={1--9},
  year={2018},
  organization={IEEE}
}

@Article{Andersson2019,
  author = {Joel A E Andersson and Joris Gillis and Greg Horn
            and James B Rawlings and Moritz Diehl},
  title = {{CasADi}: A software framework for nonlinear optimization and optimal control},
  journal = {Mathematical Programming Computation},
  volume = {11},
  number = {1},
  pages = {1--36},
  year = {2019},
  publisher = {Springer},
  doi = {10.1007/s12532-018-0139-4}
}

@article{korda2018linear,
  title={Linear predictors for nonlinear dynamical systems: {Koopman} operator meets model predictive control},
  author={Korda, Milan and Mezi{\'c}, Igor},
  journal={Automatica},
  volume={93},
  pages={149--160},
  year={2018},
  publisher={Elsevier}
}

@inproceedings{zeng2021safety,
  title={Safety-critical model predictive control with discrete-time control barrier function},
  author={Zeng, Jun and Zhang, Bike and Sreenath, Koushil},
  booktitle={2021 American Control Conference (ACC)},
  pages={3882--3889},
  year={2021},
  organization={IEEE}
}

@misc{unitree_rl_gym,
  author       = {Unitree Robotics},
  title        = {{unitree\_rl\_gym: Reinforcement Learning implementation for Unitree robots (Go2, H1, H1\_2, G1)}},
  howpublished = {\url{https://github.com/unitreerobotics/unitree_rl_gym}},
  year         = {2024},
  note         = {[Online]. Accessed: Mar. 3, 2026},
}

@misc{unitree_g1,
  author       = {Unitree Robotics},
  title        = {{Unitree G1 Humanoid Robot}},
  howpublished = {\url{https://www.unitree.com/g1/}},
  year         = {2024},
  note         = {[Online]. Accessed: Mar. 3, 2026},
}

@article{ben2025homie,
  title={{HOMIE}: Humanoid loco-manipulation with isomorphic exoskeleton cockpit},
  author={Qingwei Ben and Feiyu Jia and Jia Zeng and Junting Dong and Dahua Lin and Jiangmiao Pang},
  journal={arXiv preprint arXiv:2502.13013},
  year={2025}
}

@misc{guo2026surprisingeffectivenessspectralclipping,
  title={On the Surprising Effectiveness of Spectral Clipping in Learning Stable Linear and Latent-Linear Dynamical Systems},
  author={Hanyao Guo and Yunhai Han and Harish Ravichandar},
  year={2026},
  eprint={2412.01168},
  archivePrefix={arXiv},
  primaryClass={cs.RO},
  url={https://arxiv.org/abs/2412.01168},
}

\end{document}